\documentclass[letterpaper, 10 pt, conference]{ieeeconf}  % Comment this line out if you need a4paper

\IEEEoverridecommandlockouts                              % This command is only needed if 
\usepackage{amsmath} % assumes amsmath package installed
\usepackage{amssymb}  % assumes amsmath package installed
\usepackage{multirow}
\usepackage{graphicx}
\usepackage{subfigure}
\usepackage{float}
\usepackage{algorithm}
\usepackage{placeins}
\RequirePackage{algorithm}
\RequirePackage{algorithmic}
\RequirePackage{color}
\usepackage{colortbl}

\usepackage{dblfloatfix}
\usepackage [english]{babel}
\usepackage [autostyle, english = american]{csquotes}
\MakeOuterQuote{"}

\title{\LARGE \bf
Depthwise Multiception Convolution for Reducing Network Parameters without Sacrificing Accuracy
}

\author{Guoqing Bao, Manuel B. Graeber and Xiuying Wang% <-this % stops a space

\thanks{Guoqing Bao is with School of Computer Science, The University of Sydney, Camperdown NSW 2006, Australia (e-mail: {\tt\small guoqing.bao@sydney.edu.au}).}%
\thanks{Manuel B. Graeber is with the Ken Parker Brain Tumor Research Laboratories, Brain and Mind Centre, Faculty of Medicine and Health, The University of Sydney, Sydney, NSW, 2006, Australia ({e-mail: \tt\small manuel.graeber@sydney.edu.au}).}%
\thanks{Xiuying Wang is with School of Computer Science, The University of Sydney, Camperdown NSW 2006, Australia (Correspondence to {\tt\small xiu.wang@sydney.edu.au}).}%
}

\begin{document}

\maketitle
\thispagestyle{empty}
\pagestyle{empty}

%%%%%%%%%%%%%%%%%%%%%%%%%%%%%%%%%%%%%%%%%%%%%%%%%%%%%%%%%%%%%%%%%%%%%%%%%%%%%%%%
\begin{abstract}

Deep convolutional neural networks have been proven successful in multiple benchmark challenges in recent years. However, the performance improvements are heavily reliant on increasingly complex network architecture and a high number of parameters, which require ever increasing amounts of storage and memory capacity. Depthwise separable convolution (DSConv) can effectively reduce the number of required parameters through decoupling standard convolution into spatial and cross-channel convolution steps. However, the method causes a degradation of accuracy. To address this problem, we present depthwise multiception convolution, termed Multiception, which introduces layer-wise multiscale kernels to learn multiscale representations of all individual input channels simultaneously. We have carried out the experiment on four benchmark datasets, i.e. Cifar-10, Cifar-100, STL-10 and ImageNet32x32, using five popular CNN models, Multiception achieved accuracy promotion in all models and demonstrated higher accuracy performance compared to related works. Meanwhile, Multiception significantly reduces the number of parameters of standard convolution-based models by 32.48\% on average while still preserving accuracy. 

\end{abstract}

%%%%%%%%%%%%%%%%%%%%%%%%%%%%%%%%%%%%%%%%%%%%%%%%%%%%%%%%%%%%%%%%%%%%%%%%%%%%%%%%
\section{\textsc{Introduction}}

Convolutional neural networks [1, 2] and their successors especially AlexNet [3] have opened up a new avenue for research in computer vision. A number of highly efficient deep convolutional neural networks (DCNNs) have been designed in recent years including VGG [4], Inception [5-7] and ResNet [8]. This has become possible through carefully crafted network architectures and innovative optimization techniques such as ReLU [9], Dropout [10], Batch Normalization [6] and residual connections [8] to mitigate overfitting and to address the problem of the vanishing gradient.

The introduction of ResNet, i.e. deep residual networks, where “shortcut connections” are utilized to skip one or more layers, which eliminate singularities and alleviate the learning slow-down during network training [11], represents a milestone for training very deep networks (hundreds and even thousands of layers deep) while maintaining performance at the same time. Very sophisticated network architectures have been enabled by ResNet and they yield stunning performance. For example, Shake-Shake [12] introduced regularization for network internal representations through stochastically “blending” tensors from two residual network branches to improve generalization; Inception-ResNet [13] combined residual connections with the inception architecture in order to speed up network training and to preserve model accuracy; and SENet [14] introduced the Squeeze-and-Excitation block to allow network recalibration and to emphasize informative features.

However, the performance improvements listed above are heavily reliant on increasingly complex network architectures and a high number of parameters that are used, which in turn require a large amount of storage and consume lots of memory. Consequently, training, inference, as well as the deployment of conventional and more advanced DCNNs, have become more complicated and expensive, limiting their usage especially on mobile devices, where GPU memory resources are usually restricted.

One strategy to overcome memory restrictions is to reduce the number of network parameters via depthwise separable convolution (also dubbed DSConv), an idea derived from the concepts of “group convolution” in AlexNet [3] and separable convolution [15], respectively. DSConv decouples standard convolution into two consecutive steps: the method first performs spatial convolution on individual input channels and then utilizes 1 \textit{x} 1 kernels to conduct pointwise cross-channel convolution. DSConv was adopted by MobileNet V1 [16] and MobileNet V2 [17] for minimizing the computational density of the DCNNs. In addition, the order of the two convolution steps is reversed to form a modified version of DSConv. A new network architecture named Xception [18], resulted from this modification, achieves superior performance compared to the previous Inception architectures. Recently, Zhang \textit{et al}. introduced channel shuffle to DSConv and built a ShuffleNet architecture to improve network generalization [19].

Compared with traditional convolutional methods, DSConv is characterized by lower computational costs as well as usage of fewer parameters but suffers from a heavy degradation of accuracy [20-22] (which could also be shown in our present study). In order to address this problem, Tan \& Le \textit{et al}. recently proposed a MixConv which introduces multiscale kernels to DSConv [23]. However, MixConv is specifically designed for MixNet instead of general-purpose usage and the number of parameters is not significantly reduced compared to standard convolution. Different from the existing approach, we are introducing depthwise multiception convolution, which is characterized by two properties: 1) simultaneous learning of channel-wise multiscale representations, and 2) configurable and flexible kernel size arrangements. The proposed Multiception is intended to serve as a third option besides standard convolution and DSConv. As demonstrated in this study, Multiception reliably outperforms DSConv and MixConv in various popular network architectures while reducing the number of parameters required in standard convolution-based models by 32.48\% on average without sacrificing accuracy.

\section{\textsc{Related works}}

\subsection{Methods for model compression and parameter reduction}

Several techniques have been proposed for compressing network models and reducing the number of network parameters. They can be roughly separated into four major categories: network quantization, pruning, low-rank factorization and compact architecture. 

As the extreme version of quantization, network binarization is based on 1-bit weights instead of 32-bit float values for network training [24]. To counterbalance the performance degradation of the binary network, n-bit based quantization methods have been proposed. For example, Gupta and colleagues have shown that DCNNs can be effectively trained using 16-bit wide fixed-point representation when combined with stochastic rounding [25]. 

In addition to quantization, removing redundant network weights (or network pruning) was first suggested by LeCun et al. [26] as another promising strategy for network compression. Subsequent studies have utilized the characteristics of weight sparsity for pruning DCNN models. For example, HashedNets [27] employed low-cost hash functions to group weights into random hash buckets with a single parameter value shared between weights within the same hash bucket. Similarly, Mao \textit{et al}. [28] explored coarse-grained sparsity and showed that it can create regular sparsity patterns and that coarse-grained sparsity based pruning can achieve better compression ratios and accuracy than fine-grained sparsity and unstructured pruning. Another study [29] combined pruning, quantization and Huffman coding to achieve greater compression ratios. However, pruning usually requires fine-tuning of the parameters through manually adjusting the sensitivity for the different neural network layers, thus, the method is typically used for reducing the size of pre-trained models.

Low-rank factorization has been suggested by numerous researchers as an alternative way for reducing the number of network parameters in addition to quantization and pruning. For example, Jaderberg \textit{et al}. [30] adopted a basis of low-rank  (rank-1) filters in the spatial domain to exploit cross-channel redundancy. In contrast, Denton and his colleagues have exploited redundancy through obtaining appropriate low-rank (rank-k) approximation for each convolutional layer [31]. Unfortunately, decomposition operations in low-rank factorization require intensive computation.

In order to further address the aforementioned technical constraints, recent investigations have been focused on convolutional filters and compacting network architecture [32, 33]. A typical example is the depthwise separable convolution mentioned earlier. The DSConv and its variants have been adopted by MobileNet V1 [16], MobileNet V2 [17], Xception [18] and ShuffleNet [19] proving to be an effective way for reducing the number of network parameters and cutting computational costs. The implementation of DSConv will be discussed in section 2.2. Importantly, DSConv or DSConv-based network architectures could be combined with existing quantization and pruning methods for additional network compression.

\subsection{The mechanism of DSConv for network parameter reduction}

\begin{figure}[ht]
\begin{center}
\centerline{\includegraphics[width=\columnwidth]{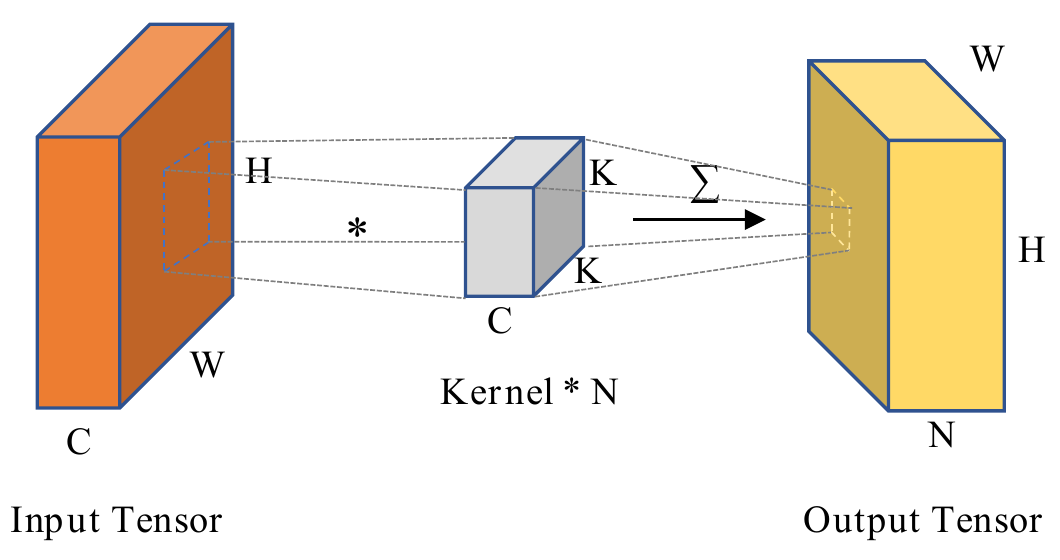}}
\vskip -0.1in
\caption{Standard convolutional operation.}
\label{fig1}
\end{center}
\vskip -0.3in
\end{figure}

Assume a convolutional layer $i$ has input and output tensors with the shape of $<H, W, C>$ and $<H, W, N>$, respectively, where $H$ and $W$ are spatial dimensions; $C$ and $N$ are input and output channel dimensions; $K$ is the kernel size. The convolutional layer $i$ with standard convolution, as illustrated in Fig 1, takes the number of parameters of $\omega_i$ and computations of $\theta_i$ as calculated below: 

\begin{equation}
    \omega_i=C*K^{2}*N
\end{equation}
\begin{equation}
    \theta_i=HWC*K^{2}*N
\end{equation}

In contrast, as suggested by a now classical paper, Krizhevsky 2012 [3], channels for the input tensor can be grouped and the convolution can be independently performed on each group, thus reducing the number of computations to $HWC * K^2 * N/G$, where $G$ is the number of groups. The most stringent version of the grouped convolution approach is called depthwise convolution [18] with $G$ equal to the number of output channels, $N$. Therefore, it only takes $HWC * K^2$ computations and $C * K^2$ parameters.  In order to obtain the appropriate number of output channels, a pointwise convolution which utilizes an $1\times1$ kernel size is performed on the products of depthwise convolution, requiring only $HWC * N$ computations and $C * N$ parameters. Thus, the layer $i$ with DSConv utilizes the following number of parameters and computations:

\begin{equation}
    \omega_i^{'}=C*(K^{2}+N)
\end{equation}
\begin{equation}
    \theta_i^{'}=HWC*(K^{2}+N)
\end{equation}

Therefore, DSConv only requires the following ratio ($\alpha$) of computational operations and the number of parameters compared to standard convolution:  

\begin{equation}
     \alpha=(K^2+N)/(K^{2}*N)
\end{equation}

\section{\textsc{Multiception Convolution}}

Inspired by DSConv and MixConv, we now propose a simple but effective convolutional method, named depthwise multiception convolution (or Multiception in short), which simultaneously performs depthwise convolution using kernels of varying sizes (each kernel size applied on all individual input channels) and concatenates the “multiscale” depthwise convolution outputs to provide input for pointwise convolution as illustrated in Fig 2. Multiception convolution empowers DSConv and enables the network to simultaneously learn higher-level multi-representations of the individual input channels instead of one state at a time. The Multiception convolution will be downgraded and has the same effects as DSConv if only fixed kernel sizes are utilized.

\begin{figure}[ht]
% \vskip 0.2in
\begin{center}
\centerline{\includegraphics[width=\columnwidth]{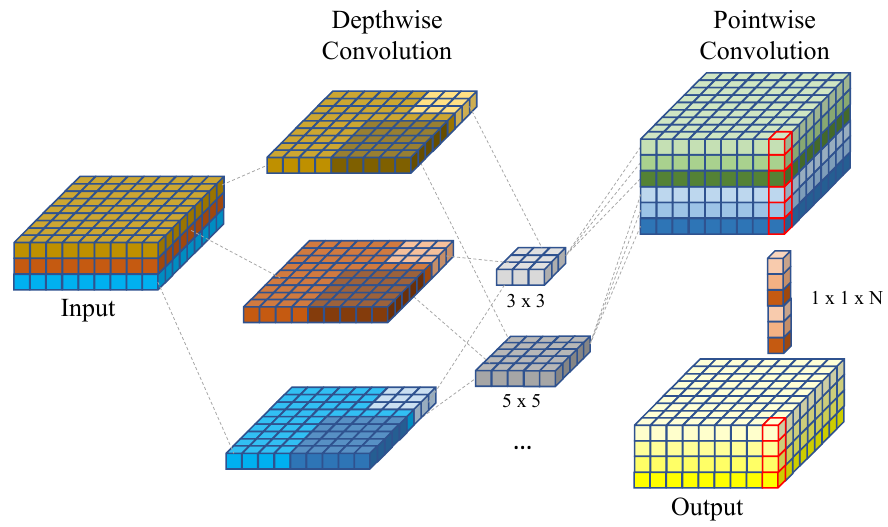}}
\vskip -0.1in
\caption{Multiception convolution with configurable kernels. Padding was used depending on the kernel size. Batch normalization and ReLU activation are not shown here.}
\label{fig2}
\end{center}
\vskip -0.1in
\end{figure}

As mentioned earlier, MixConv also harnesses multiscale kernels. However, MixConv partitions channels into different groups and applies different kernel sizes to different groups, whereas Multiception applied each selected kernel size to all channels, subsequently concatenates channel outputs and then performs pointwise convolution. In theory, Multiception captures higher level representations of all channels from different perspectives (channel-wise multi-representations), whereas, MixConv captures different representations of parts of channels (subchannel-wise multi-representations) because each group of subchannels is assigned to a fixed kernel size.

Multiception convolution can be configured to simultaneously utilize kernels of different sizes, i.e. $3\times3$, $5\times5$ and $7\times7$. Kernels that are bigger capture larger receptive fields and thus larger patterns especially in the first few layers [34], whereas small filters are more sensitive for local features. The number of kernels and the kernel size that are selected in each layer depend on their relative location in the network (layer-based kernel size arrangement). In the present setting, the first 1/3 layers harnessed all three kernel sizes and the last 1/3 convolution layers only utilized the $3\times3$ kernel (where the convolutions are downgraded to DSConv). Kernel sizes exceeding $7\times7$ were not included in view of their high parameter usage and low computational efficiency.  Even more important, Multiception convolution is very flexible and can be used interchangeably with other convolutional modules within a single neural network. The PyTorch pseudocode of multiception convolution is shown in Algorithm ~\ref{alg}.

\begin{algorithm}[ht]
   \caption{Multiception Convolution}
   \label{alg}
\begin{algorithmic}
   \STATE {\bfseries Input:} input tensor $in$, input channels $in_c$, output channels $out_c$, kernel sizes $kernels$
    \STATE {\bfseries Output:} output tensor $out$
\STATE $dsConvs = []$
\STATE $paddingDict = \{1:0, 3:1, 5:2, 7:3\}$
\STATE $groups = in_c$
\STATE {\color{blue} /* Multiscale depthwise convolution} 
\FOR{$kernel$ {\bfseries in} $kernels$}
    \STATE $padding = dict[kernel]$ 
    \STATE $sep = \texttt{Conv2d}(in,in_c,in_c, kernel, padding, groups)$
    \STATE $dsConvs.\texttt{append}(sep)$
\ENDFOR
\STATE $kSize, padding, groups, dim = 1, 0, 1, 1$
\STATE {\color{blue} /* Concatenate multiscale depthwise convolution output} 
\STATE $\hat{in_c} = in_c * \texttt{len}(kernels)$
\STATE $out = \texttt{Concatenate}(dsConvs, dim)$
\STATE $out = \texttt{BatchNorm2d}(out,\hat{in_c})$
\STATE {\color{blue} /* Pointwise convolution on concatenated output} 
\STATE $out = \texttt{Conv2d}(out, \hat{in_c}, out_c, kSize, padding, groups)$
\STATE $out = \texttt{BatchNorm2d}(out,out_c)$ \\
\STATE \textbf{return} $out$
\end{algorithmic}
\end{algorithm}

Even though multiple kernels are being used simultaneously, we are able to show that Multiception significantly reduces the required number of network parameters if the number of output channel reaches some extent (i.e. 16) and kernel sizes are properly chosen in each layer, because the maximum parameter usage in each layer $i$ is:

\begin{equation}
\hat{\omega}_i=3C*(\sum_{j=1}^{3} K{j}^{2}+N)
\end{equation}

where $K_j$ is the size of kernel $j$; $C$ and $N$  represent the number of input and output channels, respectively.

\begin{figure*}[!t]
% \vskip 0.2in
\begin{center}
\centerline{\includegraphics[width=\textwidth]{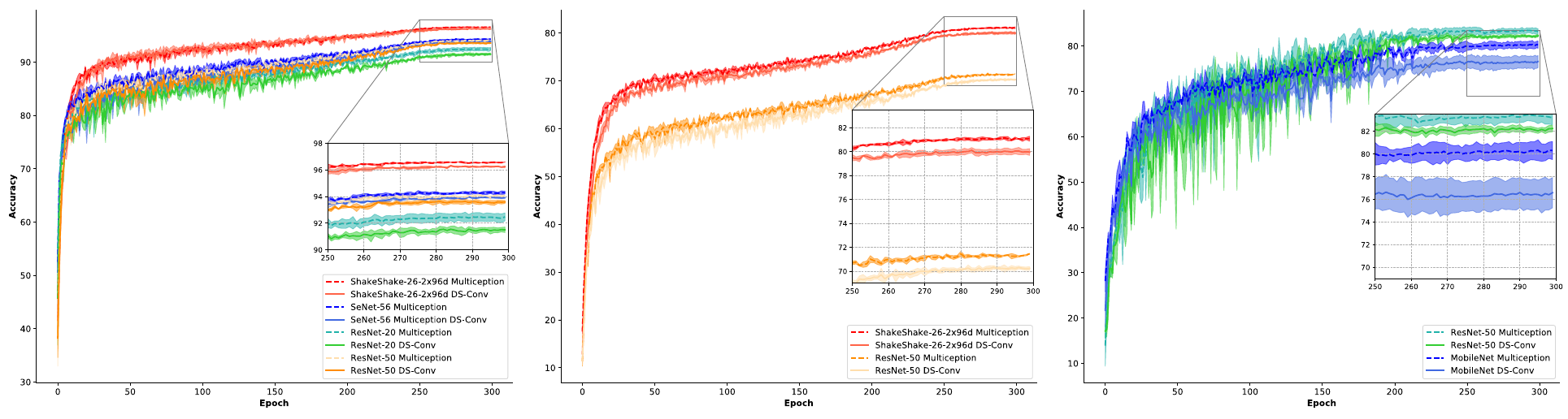}}
% \vskip -0.1in
\caption{Test accuracy comparison between Multiception and DSConv on Cifar-10 (left), Cifar-100 (middle) and STL-10 (right).}
\label{fig3}
\end{center}
\vskip -0.3in
\end{figure*}

The number of parameters used with Multiception convolution is still lower than standard convolution $C*K^{2}*N$. For example, given a network consisting of three convolutional layers, a number of input channels $c$ and the number of output channels varying between $n$, $2n$ and $4n$. A neural network with standard convolution and kernel size of $k$ would require a total of $(10n+c)*k^2*n \approx k^2*n^2$ parameters based on the calculation for each layer:

\[
\omega_1=c*n*k^2
\]
\[
\omega_2=2n^2*k^2
\]
\[
\omega_3=8n^2*k^2
\]

but the same neural network with Multiception convolution utilize approximately of $k^2*n+n^2$ parameters:

\[
\hat{\omega}_1=3C*(\sum_{j=1}^{3} K{j}^{2}+n)
\]
\[
\hat{\omega}_2=3n*(\sum_{j=1}^{2} K{j}^{2}+2n)
\]
\[
\hat{\omega}_3=6n*(k^2+4n)
\]

thus, requiring a substantially lower number of parameters when the output channel $n$  is significantly larger than the kernel size $k$.  

In the experimental section, we will show that Multiception convolution achieves better performance than DSConv and MixConv while the number of parameters used in the network remains significantly smaller compared to standard convolution.

\section{\textsc{Experiments}}

The experiments were performed on four public benchmark datasets, i.e. Cifar-10, Cifar-100 [35], STL-10 [36] and ImageNet32x32 [37, 38]. Cifar-10 and Cifar-100 consist of 10 and 100 classes and each dataset has a total of 60,000 images ($32\times32$ pixels). Both datasets were split into 50,000 images for training and 10,000 images for testing. The STL-10 consists of 10 classes of $96\times96$ pixels images with a total of 5,000 images for training and 8,000 for testing. Downsampled ImageNet forms part of ImageNet benchmark program, which poses a greater challenge since it contains all images as the original ImageNet ($224\times224$ pixels) but comes with much lower image resolutions, i.e. $64\times64$, $32\times32$ and $16\times16$. In the present study, we adopted the $32\times32$ version, dubbed ImageNet32x32, to evaluate our proposed convolutional method on a large-scale dataset.

State-of-the-art convolutional neural networks including ResNet, SeNet, ShakeNet (Shake-Shake) and MobileNet were adopted as baseline models to evaluate the effectiveness of Multiception convolution in comparison to standard convolution, DSConv and MixConv. We applied the standard training parameters, preprocessing and augmentation procedures as reported in the original publications [8, 12, 14, 17]. Specifically, we used an SGD optimizer with Nesterov momentum of 0.9, weight decay of 2E-4, 300 epochs (50 epochs on ImageNet32x32) and a batch size of 128 for Cifar and ImageNet datasets (64 for STL-10). We further used the cosine learning rate scheduler [39, 40] for training all models and the learning rate was initialized with the maximum of 0.1 and gradually decreased to a minimum of 5E-5. Since the ImageNet dataset has over 1.2 million training images, we did not perform any augmentation on it. For other datasets, we performed standard augmentations, including padding (4 pixels on each side), random crop and horizontal flip. All image inputs were normalized using means and standard deviations of three RGB channels. To make an unbiased comparison, all four convolutional methods were evaluated under consistent settings. We ran all models three times on a single NVIDIA 2080 Ti GPU and report the average top-1 test error in the last epoch.

\subsection{Classification results on Cifar-10 and Cifar-100} 

\begin{table}[ht]
\caption{Comparison of top-1 test errors (\%) on Cifar-10 and Cifar-100 datasets (average of 3 runs in last epoch).}
\centering
\arrayrulecolor{black}
\label{table1}
\resizebox{\columnwidth}{!}{%
\renewcommand{\arraystretch}{1.2} % default is 1.0
\begin{tabular}{|l|l|c|c|c|} 
\cline{1-2}\arrayrulecolor{black}\cline{3-3}\arrayrulecolor{black}\cline{4-5}
Datasets                   & Models            & DSConv & MixConv & Multiception      \\ 
\hline
\multirow{4}{*}{Cifar-10}  & ResNet-20         & 8.50   & 8.47    & \textbf{ 7.53 }   \\
                           & ResNet-50         & 6.41   & 7.16    & \textbf{ 5.94 }   \\
                           & SeNet-56          & 6.10   & 6.88    & \textbf{ 5.71 }   \\
                           & ShakeNet-26-2x96d & 3.77   & 4.38    & \textbf{ 3.42 }   \\ 
\hline
\multirow{2}{*}{Cifar-100} & ResNet-50         & 29.77  & 30.36   & \textbf{ 28.54 }  \\
                           & ShakeNet-26-2x96d & 20.03  & 20.76   & \textbf{ 18.92 }  \\
\hline
\end{tabular}
}
\end{table}

The experiment was first conducted on Cifar datasets to evaluate the effectiveness of Multiception in comparison to the closely related works (i.e. DSConv and MixConv; same kernel sizes, including $3\times3$, $5\times5$ and $7\times7$, also used for MixConv). The experimental results showed that Multiception convolution steadily outperformed the other competing methods in different network models, for example, it achieved average of 0.56\% and 1.17\% higher accuracies on Cifar-10 and Cifar-100 compared to DSConv (Table 1, and left two panels in Fig 3); and average of 1.1\%  and 1.83\% performance gains compared to MixConv (Table 1).

\subsection{Classification results on STL-10} 
ResNet-50 is widely used as the standard model for performance measurement on a variety of benchmark datasets. In addition to ResNet-50, MobileNet is more suitable for images of larger sizes, i.e. greater than $64\times64$ pixels. We have therefore conducted the experiment on STL-10 using ResNet-50 and MobileNet. Unfortunately, MixConv is not compatible with MobileNet, so the performance of MixConv on MobileNet cannot be reported. As shown in Table 2, Multiception convolution yielded higher performance compared to DSConv, especially in MobileNet (3.72\% accuracy promotion), and it also outperformed MixConv with 5.08\% promotion in ResNet-50. 
\begin{table}[ht]
\caption{Comparison of top-1 test errors (\%) on STL-10 (average of 3 runs, last epoch).}
\centering
\label{table2}
\resizebox{\columnwidth}{!}{%
\renewcommand{\arraystretch}{1.2} % default is 1.0
\begin{tabular}{|l|c|c|c|} 
\hline
Models       & DSConv & MixConv & Multiception      \\ 
\hline
ResNet-50    & 17.76  & 21.89   & \textbf{16.81}  \\ 
\hline
MobileNet-V2 & 23.42  & -       & \textbf{19.70}  \\
\hline
\end{tabular}
}
\end{table}

\subsection{Classification results on ImageNet32x32}
\begin{table}[ht]
\vskip -0.1in
\caption{Comparison of top-1 test errors (\%) on ImageNet32x32 without data augmentation (average of 3 runs, last epoch).}
\centering
\label{table3}
\resizebox{\columnwidth}{!}{%
\renewcommand{\arraystretch}{1.2} % default is 1.0
\begin{tabular}{|l|c|c|c|}
\hline
Models    & DSConv & MixConv & Multiception \\ \hline
ResNet-50 & 63.63  &   65.32      & \textbf{62.37}        \\ \hline
\end{tabular}
}
% \vskip -0.1in
\end{table}
To evaluate the effectiveness of Multiception on large-scale datasets, we conducted experiment on ImageNet32x32 using the standard ResNet-50 model without data augmentation. As shown in Table 3, Multiception surpassed both DSConv (with 1.26\% higher accuracy) and MixConv (with 3\% higher accuracy). It is worth noting that even higher performance on ImageNet32x32 could be achievable with more complex models, proper data augmentation and hyperparameter optimization, but this is beyond the scope of this study.

\subsection{Ablation study}
\begin{table}[ht]
\vskip -0.1in
\caption{Ablation study of kernel sizes used in Multiception}
\centering
\label{table4}
\resizebox{\columnwidth}{!}{%
\renewcommand{\arraystretch}{1.2} % default is 1.0
\arrayrulecolor{black}
\begin{tabular}{!{\color{black}\vrule}l|l!{\color{black}\vrule}l!{\color{black}\vrule}l!{\color{black}\vrule}l!{\color{black}\vrule}} 
\hline
Kernels                                                                         & Cifar-10 & Cifar-100 & STL-10 & ImageNet32x32  \\ 
\hline
3x3 \& 5x5                                                                        & 6.18     & 28.94     & 17.57  & \textbf{61.97}          \\ 
\arrayrulecolor{black}\hline
3x3 \& 7x7                                                                        & 5.91     & \textbf{28.54}     & 18.76  & 62.29              \\ 
\hline
5x5 \& 7x7                                                                        & \textbf{5.85}     & 29.35     & 19.59  & 62.05              \\ 
\arrayrulecolor{black}\hline
\begin{tabular}[c]{@{}l@{}} 3x3 \& 5x5 \& 7x7 \\(Present Setting) \end{tabular} & 5.94     & \textbf{28.54}     & \textbf{16.81}  & 62.37          \\
\hline
\end{tabular}
}
% \vskip -0.1in
\end{table}

As shown in Table 4, Multiception with multi-scale kernels ($3\times3$, $5\times5$ and $7\times7$) achieved best overall performance compared to two combined kernels. Multiception with $3\times3$ and $5\times5$ kernels achieved highest performance on ImageNet32x32 but lower on the rest of the datasets. In comparison, Multiception with $5\times5$ and $7\times7$ kernels yielded worst performance on STL-10, which denotes that $3\times3$ kernel is still necessary in the first few convolutional layers even though input image size is relatively large.

\subsection{Comparison with standard convolution}
\begin{table*}[!t]
\caption{Comparison of top-1 test errors (\%) between standard and Multiception convolution (average of 3 runs, last epoch).}
\vskip -0.1in
\centering
\label{table5}
\resizebox{\textwidth}{!}{%
\arrayrulecolor{black}
\renewcommand{\arraystretch}{1.2} % default is 1.0
\begin{tabular}{|l|l|cl|cl|c|c|} 
\hline
\multirow{2}{*}{Dataset}   & \multirow{2}{*}{Model}  & \multicolumn{2}{c|}{Standard Conv} & \multicolumn{2}{c|}{Multiception Conv} & \multirow{2}{*}{Error Diff.} & \multirow{2}{*}{Param Diff.}  \\ 
\cline{3-6}
                           &                         & Top-1 Err. (\%) & Params.                               & Top-1 Err. (\%) & Params.                                   &                              &                               \\ 
\hline
\multirow{4}{*}{Cifar-10}  & ResNet-20               & 7.17            & 0.2724M                               & 7.53            & 0.1899M                                   & 0.36\%                       & -30.30\%                      \\
                           & ResNet-50               & 5.75            & 0.7586M                               & 5.94            & 0.5094M                                   & 0.19\%                       & -32.85\%                      \\
                           & SeNet-56                & 5.69            & 0.8679M                               & 5.71            & 0.5886M                                   & 0.02\%                       & -32.18\%                      \\
                           & ShakeNet-26-2x96d       & 3.35            & 26.78M                                & 3.42            & 17.57M                                    & 0.07\%                       & -34.39\%                      \\ 
\hline
\multirow{2}{*}{Cifar-100} & ResNet-50               & 27.92           & 0.7644M                               & 28.54           & 0.5153M                                   & 0.62\%                       & -32.59\%                      \\
                           & ShakeNet-26x96d         & 18.44           & 26.81M                                & 18.92           & 17.59M                                    & 0.48\%                       & -34.39\%                      \\ 
\hline
STL-10                     & ResNet-50               & 17.83           & 0.7586M                               & 16.81           & 0.5094M                                   & -1.02\%                      & -32.85\%                      \\ 
\hline
ImageNet32x32                     & ResNet-50               & 63.20           & 0.8229M                               & 62.37           & 0.5738M                                   & -0.83\%                      & -30.27\%                      \\ 
\hline
\multicolumn{2}{!{\color{black}\vrule}c|}{\textbf{Average}}   &            &        &          &     & \textbf{-0.01\%}            & \textbf{-32.48\%}           \\
\hline
\end{tabular}
}
\vskip -0.2in
\end{table*}
To evaluate the performance of Multiception convolution on network parameter reduction, we compared the performance and parameter usage of Multiception to that of the standard convolution (a convolutional method that most state-of-the-art models are based on). As shown in Table 5 (next page), models with Multiception convolution require 32.48\% fewer parameters on average compared to models with standard convolution while their accuracies remain similar. Multiception achieved slightly lower accuracies on Cifar datasets but higher performance on STL-10 and ImageNet32x32 compared to standard convolution. In the case of large models (ShakeNet-26-2x96d), we found that Multiception convolution induced a slightly greater parameter reduction ($>$34\%) in comparison to smaller models, e.g. ResNet-20 ($\sim$30\%). It is worth noting that MobileNet (for STL-10) is not included in this comparison because it was designed based on DSConv and is not suitable for an evaluation using standard convolution.

\section{\textsc{Conclusions}}

In this paper, we proposed depthwise multiception convolution to mitigate the effect of accuracy degradation introduced by DSConv. Multiception enables the network to capture higher level representations of all individual input channels from different perspectives. Compared to DSConv and MixConv, models equipped with Multiception convolution can achieve higher image classification accuracies on popular benchmark datasets. Meanwhile, the proposed convolutional method can reduce the number of network parameters in baseline models (standard convolution-based) by more than 32\% while still preserving accuracy. Combining Multiception convolution with existing quantization and pruning techniques may induce further network compaction. Importantly, Multiception has a structure that is easy to adopt across different network models.

\section{\textsc{Reproducibility}}
All datasets used in this study are publicly available. The code have been released at a public repository: https://github.com/guoqingbao/Multiception.

\end{document}